# A BERT based Sentiment Analysis and Key Entity Detection Approach for Online Financial Texts

Lingyun Zhao
School of computer science and technology
Wuhan University of Technology
Wuhan, China
124665532@whut.edu.cn

Lin Li
School of computer science and technology
Wuhan University of Technology
Wuhan, China
cathylilin@whut.edu.cn

Xinhao Zheng
School of computer science and technology
Wuhan University of Technology
Wuhan, China
248681@whut.edu.cn

*Abstract*—The emergence and rapid progress of the Internet have brought ever-increasing impact on financial domain. How to rapidly and accurately mine the key information from the massive negative financial texts has become one of the key issues for investors and decision makers. Aiming at the issue, we propose a sentiment analysis and key entity detection approach based on BERT, which is applied in online financial text mining and public opinion analysis in social media. By using pre-train model, we first study sentiment analysis, and then we consider key entity detection as a sentence matching or Machine Reading Comprehension (MRC) task in different granularity. Among them, we mainly focus on negative sentimental information. We detect the specific entity by using our approach, which is different from traditional Named Entity Recognition (NER). In addition, we also use ensemble learning to improve the performance of proposed approach. Experimental results show that the performance of our approach is generally higher than SVM, LR, NBM, and BERT for two financial sentiment analysis and key entity detection datasets.

## I. INTRODUCTION

With the rapid development of the Internet, great changes have taken place in financial domain. The development of Internet finance has brought subversion to the traditional finance, and also changed people's work and life style. Compared with traditional finance, Internet finance has the characteristics of low cost, high efficiency, wide coverage and high risk. The emerging social media, represented by Micro-blog, provides relevant financial information. Sentiment analysis and key entities of online financial texts help to understand the public's sentiment state, timely access to public opinions and attitudes, and rapidly get to the subject of information. Thus it is of great practical significance for risk control, public opinion analysis and govern-ment regulation [1].

Because of the large amount of financial information on the Internet and the high value of negative information in other areas such as risk control, we are more concerned with negative information than positive information in financial domain. In addition, key entities in financial information help us to better understand the subject of the event. Thus, it can provide more targeted and accurate guidance for the decision making and investing [2]. However, the texts of social media event have feathers of massive-sparse, dynamic-heterogeneous, obscure-vague, which increase the difficulty of sentiment analysis and key entity detection [3, 4].

The existing techniques of sentiment analysis are mainly divided into three types: the sentiment dictionaries based methods [5, 6], machine learning based models [7, 8], deep learning based models [9]. Among them, deep learning methods for sentiment analysis have become very popular. They provide automatic feature extraction and both richer representation capabilities and better performance than traditional feature based methods [10].

At present, the most commonly used entity detection method is Named Entity Recognition (NER), The existing techniques of NER are mainly divided into four types: methods based on rule and dictionaries [11], methods based on statistics [12], methods based on fusion [13], methods based on deep learning [14]. The existing methods of NER are mostly based on deep learning. These methods perform better than the traditional methods. However, all methods cannot analyze the importance of detecting entities, which means that all entities will be detected from texts. Therefore, NER cannot solve the problem of detecting key entities.

In order to solve the above problems, we combine sentiment analysis and key entity detection in a unified approach based on RoBERTa [15]. We use RoBERTa as a pre-training model for fine-tuning, and different fine-tuning methods are used to implement sentiment analysis and key entity detection. We consider sentiment analysis as a classification problem to get negative emotion information, and the key entity detection as a sentence matching or Machine Reading Comprehension (MRC) tasks. Our approach's experimental results show that RoBERTa fine-tuning model performs better than other traditional models in sentiment analysis. In addition, in our proposed approach, we match each financial entity with its text to detect the key entities of negative information. In fine-grained tasks, we can even extract the most relevant key entities according to the different financial event types based on RoBERTa. However, the above tasks cannot be solved by traditional NER.

## II. RELATED WORK

### A. Sentiment Analysis in Finance

Sentiment analysis is the task of extracting sentiments or opinions of people from written language [16]. As mentioned in the previous section, we can divide the existing techniques of sentiment analysis into three types. Financial sentiment analysis differs from general sentiment analysis not only in domain, but also the purpose. The purpose behind financial sentiment analysis is usually guessing how the markets will react with the information presented in the text [17].

It is not hard for us to know that deep learning in various domains has better performance than machine learning [28-34]. Thus, it is no exception in the domain of finance. Loughran et al. [18] presents a thorough survey of recent works on financial text analysis utilizing machine learning with "bag-of-words" approach or lexicon-based methods. Loughran et al. [19] create a dictionary of financial terms with assigned values such as "positive" or "uncertain" and measure the tone of a documents by counting words with a specific dictionary value. Kraus et al. [20] first proposed methods that used deep learning for textual financial polarity analysis. They apply an LSTM neural network to predict stock-market movements and show that those methods are more accurate than traditional machine learning approaches. Sohangir et al. [21] apply several generic neural network architectures to a StockTwits dataset, finding CNN as the best performing neural network architecture.

However, due to lack of large labeled financial datasets, it is difficult to utilize neural networks to their full potential for sentiment analysis. A more promising solution could be initializing almost the entire model with pre-trained values and fine-tuning those values with respect to the classification task. Therefore, we adopt the idea of migration learning, and use the mature pre-training language model, BERT to fine-tune it to achieve sentiment analysis.

### B. Pre-trained Language Models

Since 2018, natural language processing (NLP) has opened a new era. With Devlin et al. [22] proposing BERT by Google and its outstanding performance in 11 NLP tasks, "pre-training + fine-tuning" has become one of the commonly used methods in NLP tasks. The emergence of BERT has completely changed the relationship between pre-training generated word vectors and downstream NLP tasks. Compared to the word-level vector generated by the traditional method, BERT will train the sentence-level vector and get more information from context. The sentence-level vector is more convenient for the use of downstream NLP tasks.

Although BERT has performed well and can be used as a regular component in many NLP tasks, it ignores the integration of knowledge information into language understanding. In order to solve this problem, Zhang et al. [23] proposed a model named ERNIE. It is pre-trained by masking semantic units such as words and entity concepts, rather than masking words like BERT, which makes the representation of semantic knowledge units of the model closer to the real world. Liu et al. [15] proposed RoBERTa based on BERT. Compared to BERT, RoBERTa uses more data to train, increases the batch size, removes next predict loss, and replaces static masking with dynamic masking in the pre-training stage to further improve the performance. Google again proposed the new model ALBERT by Lan et al. [25]. Compared with BERT, ALBERT adopted a new parameter sharing mechanism, which not only improved the performance of the model, but also reduced the memory usage and improved the training speed. ALBERT alleviates the problem that it's hard to use BERT to some extent. Cui et al. [26] proposed RoBERTa-wwm-ext which surpassed BERT and ERNIE in multiple tasks. The model was improved on RoBERTa, using Whole Word Masking (WWM) [27] to expand the amount of training data. In this paper, we use RoBERTa-wwm-ext as a pre-training model. Moreover, we use it for sentiment analysis tasks and key entity detection as QA tasks or sentence matching tasks in multi-granularity.

## III. OUR APPROACH

### A. Problem Description

Social media and online news about the financial domain contain a variety of entities and its sentiments. However, there are often multiple entities in one text. Some of these entities are redundant, noisy, or irrelevant. For those who pay attention to this information, such as investors and decision makers, they are more inclined to pay attention to negative information and key entities. Therefore, people need to judge the sentiment of the information and detect which is the key entity. In order to be able to make large-scale judgments on the sentiments of online financial information, and detect key entities among them. We need to analyze the sentiments of the online financial information and use a method to select key entities.

### B. Flowchart of Our Approach

At present, there are also many researches on sentiment analysis and entity detection. However, there are few related studies in financial domain, and existing entity detection methods cannot detect key entities such as NER, let alone select the most relevant entities with tags. In order to overcome the limitations of the above methods, we propose a RoBERTa based sentiment analysis and its key entity detection approach for online financial texts as follows:

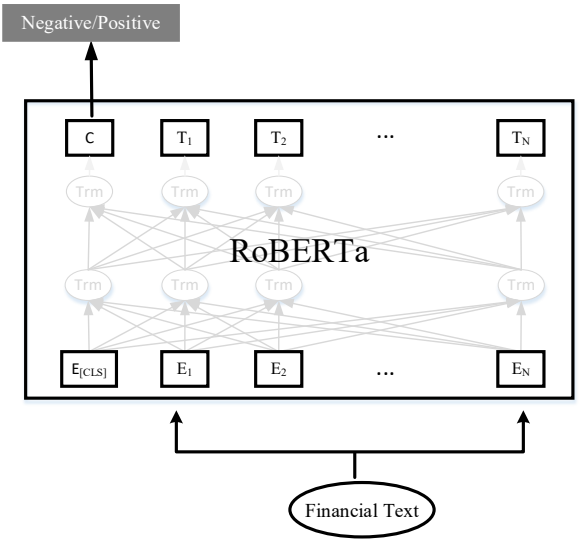

Fig. 1. Analyze the sentiment of text.

*1) Analyze the sentiment of text:* Considering that negative information pays more attention to people, we use sentiment analysis model based on RoBERTa to select negative information (Fig. 1). RoBERTa was chosen to have an identical model size as BERT and used bidirectional Transformer. The online financial text words are used as the input of the model, and sentiments are output by the models from the last layer. In this step, we fine-tuned RoBERTa to do sentiment analysis and use cross entropy loss as a loss function. It can solve the problem that it is difficult to utilize neural networks to their full potential for sentiment analysis. At the same time, we use the ensemble learning method: we train models with different random seeds, then we select ten models with higher scores to predict the results and vote to get the final answer. Besides, we also tried to re-select parameters, arrange them around multiple better parameters for training and used 10-fold cross-validation for them respectively.

*2) Get financial entity list and select key entities:* For each piece of financial data, we use NER or rule matching to get entities from the text as entity list (in some datasets, the entity list has been provided). In the coarse-grained task, we detected some key entities related to the financial text from the entity list. The key entities may be one or more. Therefore, we consider this task as a sentence matching task. As shown in Fig. 2, we enter each entity and financial text into RoBERTa. Use RoBERTa model as a sentence matching task to determine whether each entity is a key entity. The number of models is determined by the number of entities *n*. Finally, we select the entities predicted by the model as critical and form them into a list of critical entities. In this step, compared to existing entity detection approaches, the contribution of our proposed approach is to detect key entities by using sentence matching method. In addition, we found that the loss function using focal loss instead of cross entropy loss had better performance. Due to the large number of key entities in the entity list, we also tried to lower the threshold for determining key entities to obtain a higher recall rate (the default threshold is 0.5).

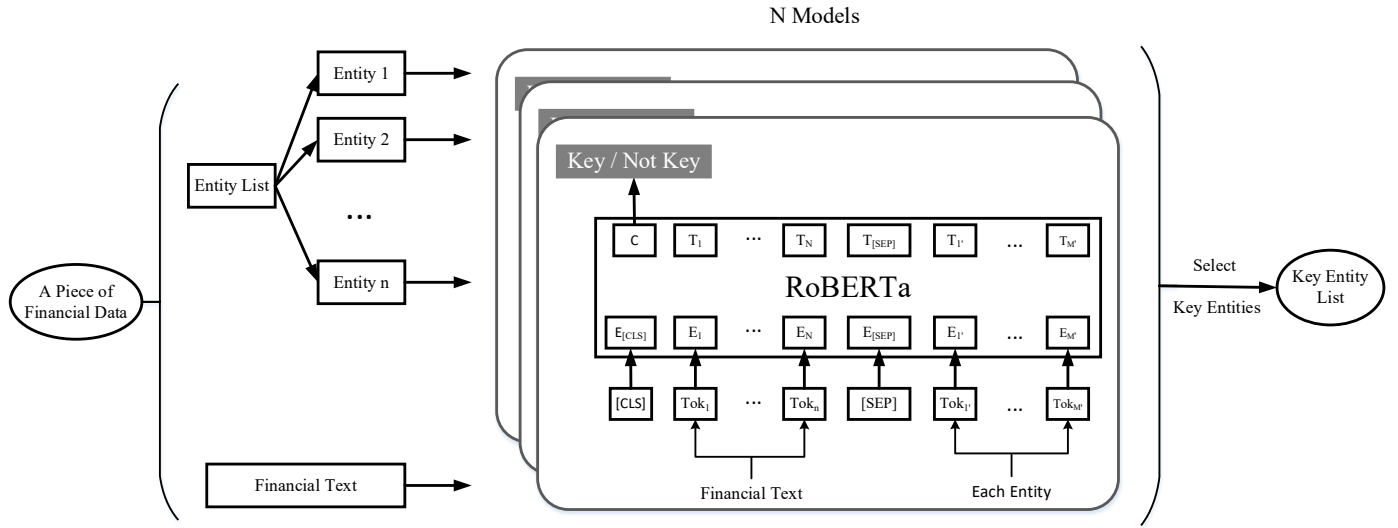

Fig. 2. Get financial entity list and select key entities.

*3) Select the key entity with the tag:* In fine-grained tasks, we need to extract the key entities that are most relevant to the given tags from the financial text. Tags can be corruption, fraud, etc. Therefore, on the basis of the previous step, we consider this task as a Machine Reading Comprehension(MRC) task (Fig. 3). We treat each financial text as an MRC article and rewrite its tag as MRC questions (for example: Which company involves *Tag*? ). Finally, we input the rewritten datasets into the MRC model of RoBERTa for training and predicting. The answers predicted by MRC model are regarded as key entities. In this step, compared to NER, the contribution of our proposed approach is to detect key entities in fine-grained tasks by using RoBERTa MRC model.

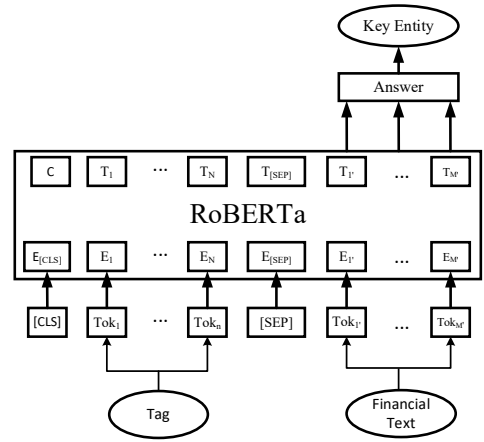

Fig. 3. Select the key entity with the tag.

## IV. Experiment Results and Analysis

### A. Data Set Description

This paper mainly studies the online financial texts. To test the effectiveness of the proposed approach in different situations. We use two sets of data in different granularity from *2019 CCF BDCI[1]* sub task as Dataset 1: Negative Financial Information And Subject Deter-mination and *2019 CCKS[2]* sub task as Dataset 2: Event Subject Extraction For Financial Field. These data sets are all online text data in financial domain. Table 1 shows the differences between the two sets of data. In *CCF BDCI* task, we need to analyze the sentiment of the text and extract the key entities of the negative information from entity lists. The task provides 17,815 pieces of data for training. In the *CCKS* task, there are 5,000 pieces of data, which is negative information, each piece of data contains a tag, and the task asks to extract the entity that is most relevant to the tag from each piece of data.

TABLE I.  DIFFERENCE BETWEEN DATASET 1 AND DATASET 2

| Data set | Sentiment | Entity List | Tag |
|---|---|---|---|
| Dataset 1 | Negative and Positive | Exist | Not Exist |
| Dataset 2 | Negative | Not Exist | Exist |

### B. Experiment Processes

There are two existing strategies for applying pre-trained language representations to downstream tasks: *feature-based* and *fine-tuning* [22]. For two sets of data, we divide them into training set and dev set by the method of 10-fold cross validation.

Experiment processes are as follows:

- Clean the data set and remove extraneous symbols, URLs, garbled characters, etc.
- Construct a new data set with each entity in the entity list and text if there is entity list in the original data set.
- Rewrite the tag as a question which is in the MRC task if there is tag in the data set.
- Train the training data set in the model to obtain the model parameters.
- Use the training parameters to predict the sentiment and key entity of the dev set.
- Compare the predicted sentiment and key entity of the dev set with its true result and assess the model by accuracy or F1.

### C. Evaluation Measures

We divided the data set into training set and dev set. In the experiment, we use Support Vector Machine (SVM), Logistics Regression (LR), Naive Bayesian Model (NBM) as baseline, BERT classifier model, and RoBERTa classifier model to train the training set for sentiment analysis and use RoBERTa sentence pair classifier model, and RoBERTa QA model to train the training set for key entity detection. Then verify the training results on the dev set to calculate the accuracy for sentiment analysis and F1 for key entity detection of each method.

*1) The accuracy rate for sentiment analysis is calculated as follows:* For the text in the dev set, if the predicted the sentiment of the text are the same as the actual sentiment of the text, the accuracy rate of the text is considered to be M/N, where M represents the correct number of predictions and N represents the total number of dev sets.

*2) The F1 score for key entity detection is calculated as follows:* For the text in the dev set, let n be the total number of texts. The Evaluation criteria is as in (6), where $TP_{ei}$ is the number of entity correctly recognized in No.*i* text, $FN_{ei}$ is the number of entity unidentified in No.*i* text, $FP_{ei}$ is the number of wrongly recognized entities in No.*i* text.

$$TP_e = \sum_{i=1}^{n} TP_{ei} \tag{1}$$

$$FP_e = \sum_{i=1}^{n} FP_{ei} \tag{2}$$

$$FN_e = \sum_{i=1}^{n} FN_{ei} \tag{3}$$

$$P_e = \frac{TP_e}{TP_e + FP_e} \tag{4}$$

$$R_e = \frac{TP_e}{TP_e + FP_e} \tag{5}$$

$$F_1^e = \frac{2P_e R_e}{P_e + R_e} \tag{6}$$

### D. Experiment Results and Quantitative Analysis

In Dataset 1's sentiment analysis, we use RoBERTa-wwm-ext fine-tuning model to predict sentiment. Similarly, we also use BERT fine-tuning model, and use RoBERTa or BERT to generate the sentence-level vector to connect the downstream model. In the experimental analysis, due to the setting of hyper-parameters has an important influence on the quality of the model, we determine hyper-parameters based on the experience of previous researchers. The results of using 10-fold cross validation for each model are shown in TABLE II. Through this table, we find that the classification of BERT fine-tuning model is generally better than BERT generating the sentence-level vector with the downstream model. RoBERTa is better than BERT in both the fine-tuning and the performance-level vector and the downstream model. In addition, we use the ensemble method [27] to improve the performance of RoBERTa to some extent.

TABLE II.  RESULT OF SENTIMENT ANALYSIS

| Model | Accuracy |
|---|---|
| $BERT_{BASE}$ to Vector + NBM | 0.87750 |
| $BERT_{BASE}$ to Vector + LR | 0.91957 |
| $BERT_{BASE}$ to Vector + SVM | 0.94034 |
| $RoBERTa_{BASE}$ to Vector + SVM | 0.94762 |

| Model | Accuracy |
|---|---|
| BERT$_{BASE}$-Single | 0.95427 |
| RoBERTa$_{BASE}$-Single | 0.96638 |
| Our Approach | **0.96774** |

In key entity detection, we use the method proposed in the previous section: extract each entity in the entity list and match it with text as a sentence to construct a new data set; if there are tags in data set, we treat the tag as a question and make it into a QA task. However, during the experiment, we found that most of the entities in the entity list are key entities, which means that the positive and negative samples are not evenly distributed in dataset 1. Therefore, we use focal-loss as a loss function and set the threshold value of sentence matching as 0.2. Finally, we use ensemble method to get the final results as shown in TABLE III and TABLE IV. Through the quantitative experimental results, we found that our proposed approach has quite good performance in both dataset 1 and dataset 2, and RoBERTa is better than BERT. Therefore, our proposed approach is effective and feasible.

TABLE III.  RESULT OF KEY ENTITY DETECTION FROM DATASET 1

| Model | F1 Score |
|---|---|
| BERT$_{BASE}$-Single | 0.94537 |
| RoBERTa$_{BASE}$-Single | 0.94953 |
| Our Approach | 0.95082 |
| Our Approach -Threshold0.2-Ensemble | 0.95140 |
| Our Approach -Threshold0.2-Ensemble-Focal-loss | **0.95258** |

TABLE IV.  RESULT OF KEY ENTITY DETECTION FROM DATASET 2

| Model | F1 Score |
|---|---|
| Dictionary matching | 0.32431 |
| BERT$_{BASE}$-QA | 0.84716 |
| Our Approach | **0.85056** |

## V. CONCLUSIONS AND FUTURE WORK

With the rapid development of Internet finance, more and more investors and decision makers need to detect key entities of negative information. However, there are few related researches in financial domain and traditional entity detection methods cannot extract key entities. In order to solve this problem, we propose a sentiment analysis and key entity detection approach. In addition, the tag is used as a constraint to the approach to further extract the key entity most relevant to the tag. The experimental results show that the performance of using the pre-training model to fine-tune is better than using the pre-training model to generate the sentence-level vector to connect to the downstream model, and the ensemble learning and the use of focal loss can improve the performance to some extent. However, there is room for improvement in our method. Due to equipment limitations, we only use the RoBEERTa-base model. If we use the RoBERTa-large model, we may get better results. The research in this paper can be used for risk control, sentiment analysis in financial domain, and helps decision makers to develop procedures or provide relevant information to investors. Furthermore, our approach is not limited to the financial sector, it can also analyze sentiment and detect key information from other domains.

---

[1] https://www.datafountain.cn/competitions/353

[2] https://www.biendata.com/competition/ccks_2019_4/